\documentclass{IOS-Book-Article}

\usepackage{mathptmx}
\usepackage{amsmath}
\usepackage{soul}\setuldepth{article}
\usepackage{graphicx}

\def\hb{\hbox to 11.5 cm{}}

\begin{document}

\pagestyle{headings}
\def\thepage{}

\begin{frontmatter}              % The preamble begins here.

%\pretitle{Pretitle}
\title{Enhancing Logical Reasoning in Large Language Models to Facilitate Legal Applications}
\author[A]{Ha-Thanh Nguyen}
\author[A]{Wachara Fungwacharakorn}
\author[A]{Ken Satoh}
\address[A]{National Institute of Informatics, Japan}
% \textit{nguyenhathanh@nii.ac.jp}
% \address[B]{Another Institution, Country}
% \email

\begin{abstract}
Language serves as a vehicle for conveying thought, enabling communication among individuals. The ability to distinguish between diverse concepts, identify fairness and injustice, and comprehend a range of legal notions fundamentally relies on logical reasoning. Large Language Models (LLMs) attempt to emulate human language understanding and generation, but their competency in logical reasoning remains limited. This paper seeks to address the philosophical question: How can we effectively teach logical reasoning to LLMs while maintaining a deep understanding of the intricate relationship between language and logic? By focusing on bolstering LLMs' capabilities in logical reasoning, we aim to expand their applicability in law and other logic-intensive disciplines. To this end, we propose a Reinforcement Learning from Logical Feedback (RLLF) approach, which serves as a potential framework for refining LLMs' reasoning capacities. Through RLLF and a revised evaluation methodology, we explore new avenues for research in this domain and contribute to the development of LLMs capable of handling complex legal reasoning tasks while acknowledging the fundamental connection between language and logic.
\end{abstract}

\end{frontmatter}

\section{Introduction}

The legal domain presents a unique set of challenges for artificial intelligence (AI) applications, as it demands high-quality reasoning capabilities, the ability to interpret complex language structures, and accurate decision-making based on legal precedence and context. While the advancement of deep learning techniques has led to the development of sophisticated language models, their current inability to exhibit reliable and consistent logical reasoning limits their applicability for critical functions such as legal advice and case analysis \cite{Rabelo_2020, Rabelo_2022, Thanh_2021, Nguyen_2022}.

Large Language Models (LLMs) like GPT-3 \cite{brown2020language} and GPT-4 \cite{openai2023gpt4} excel in various language tasks, often providing promising results for areas such as natural language understanding and generation, translation, and question answering. However, their utility in the legal domain is hindered by their weakness in processing complex logic and reasoning requirements. Recent papers highlighting the challenges faced by GPT models in logical reasoning tasks and revealing the limitations of pre-trained legal models in abductive reasoning \cite{nguyen2023negation, nguyen2023well} further emphasize the need to improve LLMs' logical reasoning capabilities for legal applications.

In response to these challenges, various techniques have been proposed to address the limitations of LLMs, such as differentiable symbolic programming \cite{zhang2023improved}, selection-inference frameworks \cite{creswell2205selection}, and graph-based reasoning methods \cite{lei2023boosting}. A comprehensive overview of reasoning techniques and benchmarks for LLMs has been provided by \cite{huang2022towards}, offering valuable insights into the current state of knowledge in this area. Furthermore, studies focusing on specific legal contexts, such as tax law \cite{nay2023large}, have demonstrated the potential for LLMs, especially when combined with effective prompting and relevant legal resources, in assisting with real-world legal tasks. However, these approaches largely overlook the potential benefits of reinforcement learning techniques that have shown considerable promise in enhancing other aspects of LLM capabilities.

In light of these observations, this work focuses on the philosophical question of how we can effectively teach logical reasoning to LLMs while maintaining a deep understanding of the intricate relationship between language and logic. To this end, we propose a Reinforcement Learning from Logical Feedback (RLLF) approach as a potential framework to refine LLMs' reasoning capacities in the context of legal applications. The suggested RLLF approach aims to improve LLMs' logical reasoning performance by focusing on logical feedback rather than relying on subjective human feedback, thus addressing the semantic challenges and inconsistencies of current models.

\section{Reinforcement Learning from Human Feedback (RLHF) and Its Limitations}

Reinforcement Learning from Human Feedback (RLHF) is an approach that leverages human judgments to guide the learning process of AI models. This approach usually comprises the following three steps:

\begin{enumerate}
    \item Generating a set of responses $\lbrace \tau_1, \dots, \tau_n \rbrace$ using a learned policy. The policy's parameters are learned via traditional reinforcement learning algorithms to maximize total reward.
    \item Selecting two segments $(\sigma_1, \sigma_2)$ from the generated responses and having a human compare and rank them in terms of performance. Human judgments are stored as a tuple $(\sigma_1, \sigma_2, \mu)$, where $\mu$ denotes the distribution of preferred segments. In each segment $\sigma_1$ and $\sigma_2$, actions $a_1^t$ and $a_2^t$ are taken at each time step $t$.
    \item Training the reward predictor using supervised learning techniques. The Bradley-Terry model is employed to estimate the reward predictor by expressing the preferred strategy through comparisons of responses. The comparisons are computed as follows: 
    $${\hat P}[\sigma_1 \succ \sigma_2] = \frac{\exp(\sum \hat{r}(\sigma_1, t, a_1^t))}{\exp(\sum \hat{r}(\sigma_1, t, a_1^t)) + \exp(\sum \hat{r}(\sigma_2, t, a_2^t))}$$ where $\succ$ denotes the preference relation. The loss function can then be written as: $\text{loss}(\hat{r}) = \sum_{(\sigma_1, \sigma_2, \mu)} \mu(1) \hat P[\sigma_1 \succ \sigma_2] + \mu(2) \hat P[\sigma_2 \succ \sigma_1]$.
\end{enumerate}

While RLHF has been successful in various applications, such as ChatGPT \cite{ouyang2022training}, it suffers from certain limitations, primarily stemming from biases present in human feedback. With a limited number of parameters, LLMs may overemphasize user satisfaction at the expense of factual accuracy and reasoning capabilities.

Human feedback may contain inherent biases, leading to suboptimal guidance during the model training process. Consequently, an LLM that focuses too heavily on user satisfaction could compromise its ability to accurately deliver facts or exhibit robust logical reasoning.

In light of these limitations, there is a need for alternative approaches that are less prone to human biases and can effectively enhance LLMs' logical reasoning capabilities, particularly in domains like law that demand both accuracy and reasoning proficiency. The proposed Reinforcement Learning from Logical Feedback (RLLF) is one such approach, specifically designed to address these concerns and ensure the enhancement of logical reasoning in LLMs without being extensively influenced by subjective human feedback.

\section{Reinforcement Learning from Logical Feedback (RLLF)}

\begin{figure}[h]
  \centering
  \includegraphics[width=0.8\textwidth]{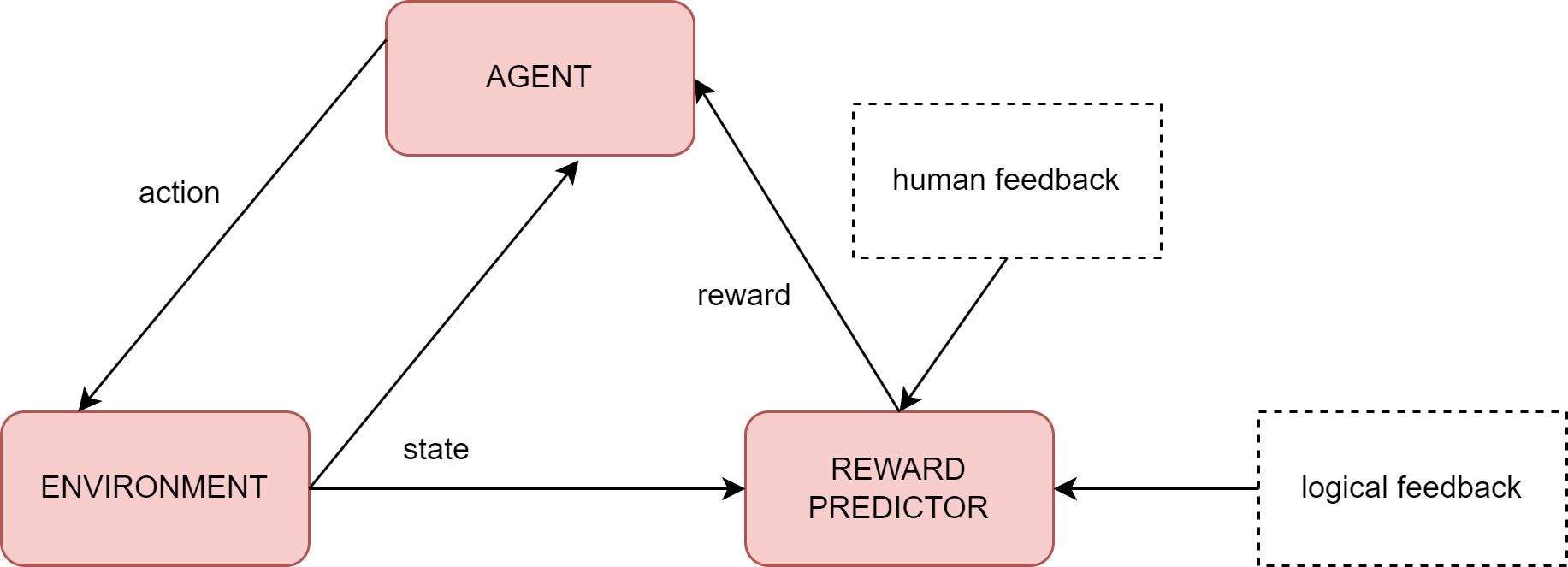}
  \caption{RLLF is the idea of allowing feedback for reinforcement learning to come not only from the user but also from the accuracy in the chain of logical reasoning.}
  \label{fig:rllf}
\end{figure}

In contrast to the reliance on human feedback in RLHF, our proposed Reinforcement Learning from Logical Feedback (RLLF) approach allows the reward predictor to be trained using logical feedback in addition to human evaluations, as illustrated in Figure \ref{fig:rllf}. This method aims to enhance LLMs' logical reasoning capabilities while minimizing the impact of human biases.

The RLLF framework comprises the following steps:

\begin{enumerate}
    \item Similar to RLHF, a set of responses $\lbrace \tau_1, \dots, \tau_n \rbrace$ is generated using a learned policy, and human feedback is collected.
        
    \item The reward predictor is trained using both human feedback and logical feedback. In addition to human feedback, the logical reasoning engine, such as a Prolog engine, is utilized as a teacher for reward predictor. During training, a suitable hyperparameter is introduced to balance the importance of user satisfaction and logical reasoning accuracy. The choice of balance depends on the specific requirements of the target application and the model's parameter constraints.
    
    \item Once trained, the reward predictor provides feedback for the reinforcement learning agent, taking into account both user satisfaction and logical reasoning accuracy. The LLM is then retrained using reinforcement learning algorithms, optimizing for the computed reward, which leads to improved logical reasoning capabilities and user satisfaction.
\end{enumerate}

Through the RLLF framework, LLMs can enhance their logical reasoning capabilities while minimizing the influence of human biases and subjective feedback. This innovative approach is particularly suitable for logic-intensive domains such as law, where the ability to reason logically and provide accurate information is essential for practical applications and decision-making processes.

Considering the limitations of RLHF in terms of biases affecting model performance and the need for improving logical reasoning in LLMs, the proposed RLLF is a promising avenue for future research. By leveraging the connections between user satisfaction and logical reasoning, we can develop LLMs that are both effective in catering to user-specific needs and proficient in providing accurate, logically sound solutions in the context of legal applications and beyond.

\section{Exploring Logical Feedback}

In this section, we explore which logical representation shall be used for logical feedback. We discuss several representation features that may affect the capabilities of capturing LLMs' logical reasoning. As researchers have explored various representations for law and logic, we categorize some of them as follows (some representations may belong to multiple categories):

\begin{enumerate}
    \item \textbf{Programming languages for general purpose}, including logic programming languages like Prolog and functional languages like Haskell.
    \item \textbf{Programming languages for the law}, for example, Proleg \cite{satoh2010proleg}, Catala \cite{catala2021}, etc.
    \item \textbf{Web-based languages}, for example, Resource Description Framework (RDF), Semantic Web Rule Language (SWRL), etc.
    \item \textbf{Controlled natural languages}, for example, Attempto Controlled English \cite{fuchs2008attempto}, Logical English \cite{kowalski2020logical}, etc.
    \item \textbf{Mathematical logic and abstract structure}, including classical logic like propositional logic, first-order logic and non-classical logic like linear temporal logic, and graph-based structure.
\end{enumerate}

\noindent We discuss several features as follows:

\begin{enumerate}

    \item \textbf{Popularity of representation}: Although LLMs' may induce less known representation by giving few examples -- which is known as \emph{few-shot learning} -- we hypothesize that popularity of representation may affect the capacities of capturing LLMs' logical reasoning. Well-known representation, such as Prolog, might be more effective to capture the LLMs' logical reasoning since LLMs are pre-trained with more examples in Prolog.
    
    \item \textbf{Complexity of representation}:  Since LLMs can induce representation with syntax errors, which could not be complied by the reason engine, representation with complex syntax may reduce  capacities of capturing LLMs' logical reasoning. For example, it is shown that LLMs can induce Prolog facts with missing closing parenthesis \cite{nguyen2022multi}.

    \item \textbf{Level of representation}: Although we do not mainly aim for human-readable representations, we hypothesize that high-level representation, such as controlled natural language, may affect the capacities of capturing LLMs' logical reasoning since LLMs are pre-trained on vast natural language datasets.
    
    \item \textbf{Frequently used environment for representation}: The environment in which the representation frequently uses may affect the ability to capture logical reasoning of LLMs. For example, if we force LLMs to induce web-based representations for law, LLMs may recall knowledge that is stored in the web-based data as relevant, while this may not happen when we force LLMs to induce other kinds of representation.

    \item \textbf{Logical value assumption}: Logical value assumptions, such as closed-world/ open-world assumptions, may lead to bias in logical reasoning engines. For example, if we focus on the legal textual entailment task, given a question Q and a relevant legal rule, the task is to determine if the legal rule entails the ``Q'' (true) or ``not Q'' (false), a Prolog engine tends to answer false because Prolog is based on closed-world assumption.

    \item \textbf{Solution determination}: The way in which the logic of the representation determines solutions is an interesting feature to consider. For example, several representations are based on answer set programming, which can generate multiple solutions. This may lead to multiple feedbacks for reinforcement learning.
    
    \item \textbf{Representation consistency}: Representation consistency is another interesting feature to consider. For example, in the legal textual entailment task, it is interesting how representations vary the capacities of LLMs to induce consistent structures for the same legal rule but different questions.
\end{enumerate}

\section{Conclusion}

This work addresses the current limitations of Large Language Models (LLMs) in logical reasoning, particularly in the legal domain. We propose the Reinforcement Learning from Logical Feedback (RLLF) approach, which offers an alternative to traditional Reinforcement Learning from Human Feedback (RLHF) methods. By incorporating logical reasoning feedback in addition to human feedback, RLLF aims to mitigate the influence of human biases and improve LLMs' logical reasoning capabilities. The RLLF framework holds promise for future research and practical applications in law and other logic-intensive disciplines, fostering the development of LLMs that effectively cater to user-specific needs while providing accurate and logically sound solutions.

\section*{Acknowledgements}
This work was supported by the AIP challenge funding related with JST, AIP Trilateral AI Research, Grant Number JPMJCR20G4.

\bibliographystyle{vancouver}
\bibliography{ref.bib}

% \begin{thebibliography}{99}

% \bibitem{r1}
% Petitti DB, Crooks VC, Buckwalter JG, Chiu V. Blood pressure levels before dementia.
% Arch Neurol. 2005 Jan;62(1):112-6.

% \bibitem{r2}
% Rice AS, Farquhar-Smith WP, Bridges D, Brooks JW. Canabinoids and pain. In: Dostorovsky JO,
% Carr DB, Koltzenburg M, editors. Proceedings of the 10th World Congress on Pain;  2002 Aug
% 17-22; San Diego, CA. Seattle (WA): IASP Press; c2003. p. 437-68.

% \end{thebibliography}
\end{document}